\documentclass[runningheads]{llncs}
\usepackage[T1]{fontenc}
\usepackage{cite}
\usepackage{soul}
\usepackage{xcolor}
\usepackage{csquotes}
\usepackage{marvosym} 
\usepackage{hyperref} 
\usepackage{amsmath, amsfonts}
\usepackage{graphicx, verbatim}

\PassOptionsToPackage{hyphens}{url}

\begin{document}
%
\title{A Hybrid Fully Convolutional CNN-Transformer Model for Inherently Interpretable Disease Detection from Retinal Fundus Images}

\titlerunning{Self-explainable CNN-Transformer model for retinal fundus disease detection}

\author{Kerol Djoumessi\inst{1}\textsuperscript{(\Letter)}\orcidID{0009-0004-1548-9758} 
\and
Samuel Ofosu Mensah\inst{1}\orcidID{0000-0002-9290-1206} \and
Philipp Berens\inst{1,2}\textsuperscript{(\Letter)}\orcidID{0000-0002-0199-4727}
}

\authorrunning{F. Author et al.}
%
\institute{Hertie Institute for AI in Brain Health, University of T\unexpanded{\"u}bingen, Germany 
\email{ \{kerol.djoumessi-donteu, philipp.berens\}@uni-tuebingen.de}\\
\and T\unexpanded{\"u}bingen AI Center, University of T\unexpanded{\"u}bingen, Germany\\
}

\authorrunning{K. Djoumessi et al.} 
 
\maketitle              
\begin{abstract}
    In many medical imaging tasks, convolutional neural networks (CNNs) efficiently extract local features hierarchically. More recently, vision transformers (ViTs) have gained popularity, using self-attention mechanisms to capture global dependencies, but lacking the inherent spatial localization of convolutions. Therefore, hybrid models combining CNNs and ViTs have been developed to combine the strengths of both architectures. However, such hybrid models are difficult to interpret, which hinders their application in medical imaging. In this work, we introduce an interpretable-by-design hybrid fully convolutional CNN-Transformer architecture for retinal disease detection. Unlike widely used post-hoc saliency methods for ViTs, our approach generates faithful and localized evidence maps that directly reflect the model’s decision process. We evaluated our method on two medical tasks focused on disease detection using color fundus images. Our model achieves state-of-the-art predictive performance compared to black-box and interpretable models and provides class-specific sparse evidence maps in a single forward pass.

    
\end{abstract}

\section{Introduction} 
   Convolutional neural networks (CNNs) are at the heart of many successful applications in medical image analysis \cite{chen2025review}, but more recently, vision transformers (ViTs) have emerged as a competitive alternative  \cite{dosovitskiy2020image}, demonstrating strong performance in medical imaging tasks \cite{azad2024advances, takahashi2024comparison}. Although CNNs are highly effective at capturing complex local patterns in images, the size of their receptive field is smaller than some disease-related lesions \cite{donteu2023sparse}. In contrast, vision transformers leverage self-attention (SA) \cite{vaswani2017attention} to capture long-range dependencies, providing a more global understanding of the image. Despite these advantages, ViTs require substantial computational resources, often demanding large-scale datasets for effective training \cite{mauricio2023comparing, takahashi2024comparison}, while also facing challenges in interpretability \cite{kashefi2023explainability}. 
    
    To address the weaknesses of both approaches, a promising alternative are hybrid CNN-Transformer architectures. Several studies have used such architectures \cite{ilyas2024hybrid, kim2025systematic, mauricio2023comparing, takahashi2024comparison}, improving performance for tasks that require combining local features with global relationships for classification. Yet, the interpretability of such hybrid approaches has remained a challenge \cite{kim2025systematic, mauricio2023comparing, takahashi2024comparison}, as they require either techniques tailored to transformer architectures or the development of novel visualization methods \cite{kim2025systematic}. To this end, either CNN-based methods have been adapted to ViTs \cite{selvaraju2017grad, bach2015pixel} or ViT-specific techniques have been proposed \cite{abnar2020quantifying, chefer2021generic, chefer2021transformer}. The most commonly used ViT-specific approach has been to visualize attention maps across layers, as these capture interactions between input regions. However, attention is not class-specific and merely illustrates relationships between input patches rather than their direct contribution to the model prediction \cite{bibal2022attention, stassin2023explainability, kashefi2023explainability}.
    Alternatively, post-hoc CNN-based methods like LRP \cite{bach2015pixel} and GradCAM \cite{selvaraju2017grad} have been successfully adapted to ViT by integrating gradients within the self-attention layers, offering class-wise explanations \cite{chefer2021transformer}. Yet, these are model-specific and struggle with hierarchical architectures like the Swin Transformer \cite{nguyen2023inspecting}. 

    Here, we propose a novel, inherently interpretable-by-design hybrid CNN-Transformer architecture for retinal fundus image classification, combining the feature extraction strengths of CNNs with the ability of ViTs to capture long-range dependencies from dual-resolution features. Dual-resolution self-attention (DRSA) allows the model to capture both fine-grained details and global context by attending to representations at two distinct spatial resolutions. Our design integrates recent advancements such as convolutional ViTs \cite{wu2021cvt}, dual-resolution self-attention \cite{ilyas2024hybrid}, and sparse explanations \cite{donteu2023sparse, djoumessi2025soft}. We evaluated our model using two backbone CNNs---ResNet and BagNet---on two clinically relevant tasks: Diabetic Retinopathy (DR) detection and Age-Related Macular Degeneration (AMD) severity classification, using publicly available color fundus image datasets.     
    Our hybrid model provides self-interpretability without sacrificing classification performance, challenging the myth of the accuracy-interpretability tradeoff \cite{rudin2019stop}. 
    It maintained predictive performance compared to both interpretable and non-interpretable state-of-the-art models while offering faithful explanations that accurately localize disease-related lesions---even under distribution shift---outperforming traditional post-hoc methods.

\section{Developing a self-explainable hybrid CNN-ViT model} 
        \label{method}

        \begin{figure}[t]
            \centering            \includegraphics[width=1.02\textwidth]{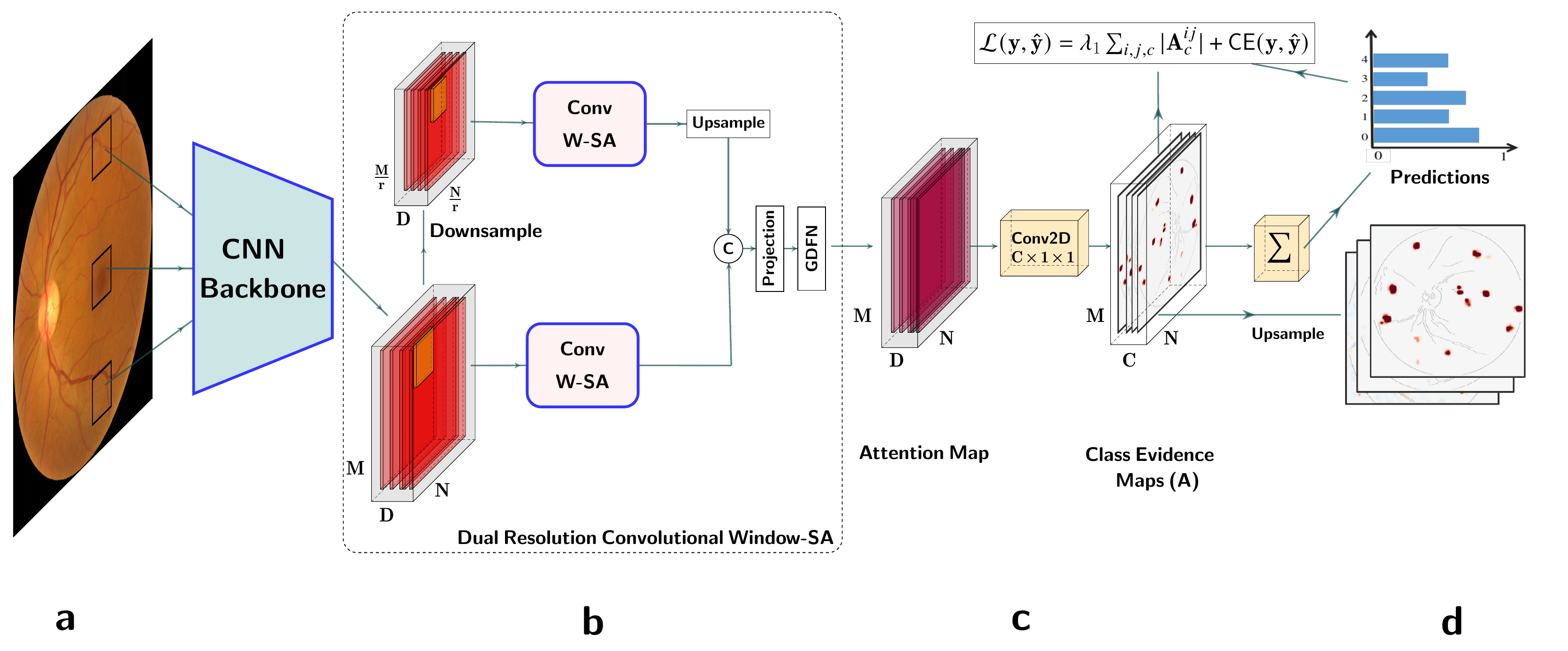} 
            \caption{\textbf{Interpretable-by-design hybrid CNN-Transformer model.} (\textbf{a}) Input image. The black patches illustrate the small receptive field of the CNN backbone (BagNet). (\textbf{b}) Two window-based SA modules are applied separately to the high and downsampled low-resolution feature maps, followed by feature fusion.
            (\textbf{c}) The high-dimensional attention map is transformed into the class evidence map $\mathbf{A}$ by applying a $1 \times 1$ convolutional classifier with $C$ kernels, where $C$ is the number of classes.
            (\textbf{d}) Spatial averaging of the class evidence maps yields predictions, whereas upsampling $\mathbf{A}$ provides explanations.} 
            \label{fig1:architecture}
            \vspace{-0.07cm}
        \end{figure}
        
        \subsection{Hybrid CNN-ViT architecture}
            \label{backbone}
            In our hybrid architecture (Fig.\,\ref{fig1:architecture}), CNN and ViT modules are used sequentially, with the output of the CNN module serving directly as the input to the transformer module. Specifically, the CNN module acted as a feature extractor, capturing local patterns. The ViT module modeled long-range dependencies between the extracted features, enhancing the model’s ability to understand broader contexts. 
            Given an input image $\textbf{X} \in \mathbb{R}^{H \times W \times C}$---where $H$, and $W$, denote height and width, and $C$ is the number of channels---the CNN backbone $f$ extracts a spatial feature representation $\mathbf{Z}=f_{\theta}(\mathbf{X}) \in \mathbb{R}^{M \times N \times D}$, where $\theta$ denotes the model parameter, $M \times N$ represents the spatial size, and $D$ is the feature dimension. We used either a ResNet50 (with a receptive field of 427×427) or a BagNet-33 (33×33) as the backbone network. Unlike the ResNet, the BagNet aggregates only local features in a bag-of-words manner \cite{brendel2018bagnets}.                  
            The transformer module (Fig.\,\ref{fig1:architecture}b) uses a dual-convolutional window self-attention (Conv-wSA) mechanism that operates on both high- and low-resolution versions of the original feature maps to produce an attention map $\mathbf{W} = g_{\phi}(\mathbf{Z}_{h}, \mathbf{Z}_{l}) \in \mathbb{R}^{M \times N \times D}$. 
            Here, $\mathbf{Z}_{h}=\mathbf{Z}$ denotes the high-resolution feature map, while $\mathbf{Z}_{l}=d(\mathbf{Z}, r) \in \mathbb{R}^{\frac{M}{r} \times \frac{N}{r} \times D}$ is the low-resolution counterpart, obtained via the downsampling function $d(.)$ with reduction factor $r$. The function $g_\phi$, parametrized by $\phi$,  jointly encompasses the parameterized projection and the Gated-Dconv Feed-Forward Network (GDFN) \cite{zamir2022restormer}. 
            The attention map produced by the transformer module maintains the spatial resolution of the input feature map.            
            The classification module (Fig.\,\ref{fig1:architecture}c) comprises a convolutional layer with $C$ kernels of size $1 \times 1$ and unit stride, producing an evidence map $\mathbf{A} = h_{\psi}(\mathbf{W}) \in \mathbb{R}^{M \times N \times C}$, where $C$ represents the number of classes and $\psi$ denotes the parameter of the classifier $h$. The final prediction $\mathbf{\hat{y}} \in \mathbb{R}^{1 \times C}$ is obtained by applying spatial average pooling to $\mathbf{A}$, followed by a softmax operation: $\mathbf{\hat{y}} = \mbox{Softmax} \big( \mbox{AvgPool}(\mathbf{A}) \big)$. This yields a $C$-dimensional probability distribution representing the likelihood of each class.

    \subsection{Learning long-range dependencies with convolutional DRSA}
        \label{drsa}        
        To learn long-range dependencies between the convolutional features, we used a transformer module with dual-resolution self-attention (DRSA) \cite{ilyas2024hybrid}, for which a convolutional layer had replaced the linear fully connected layer (FCL) \cite{wu2021cvt} as follows: $\text{SA}_h = \text{Softmax} \big( \frac{\mathbf{Q}_h \mathbf{K}^\top_h}{\alpha} \big) \mathbf{V}_h$, $\text{SA}_l = \text{Softmax} \big( \frac{\mathbf{Q}_l \mathbf{K}^\top_l}{\alpha} \big) \mathbf{V}_l$ where $\alpha$ is the scaling factor, and $\mathbf{Q}_h, \mathbf{K}_h, \mathbf{V}_h$ and $\mathbf{Q}_l, \mathbf{K}_l, \mathbf{V}_l$ are the queries, keys, and value embeddings generated for $\mathbf{Z}_h$ and $\mathbf{Z}_l$ using convolutional operations. 
        The final self-attention representation is computed as: $\text{SA}_{final} = \text{GDFN}_\delta \big(\text{Proj}_\beta(\text{SA}_h +  \text{Up}(\text{SA}_l)) \big)$, with $\mathbf{W} = \text{SA}_{final}$, where $\text{Up}(\text{SA}_l)$ denotes the upsampled version of $\text{SA}_l$. This upsampled map is aggregated with $\text{SA}_h$ and passed through a convolutional projection parametrized by $\beta$. 
        The resulting representation is subsequently refined using a GDFN parametrized by $\delta$, which enhances spatial structures while suppressing irrelevant features. This refinement ensures that only salient information contributes to the final predictions, thereby improving the generalization performance of the model.    

    \subsection{Enhancing interpretability with a sparse convolutional classifier}
        \label{explainability}
        In standard ViT and hybrid CNN-Transformer models, the classification head includes a FCL, which discards spatial information, limiting interpretability. Our architecture addressed this by preserving spatial information using convolutional operations in the self-attention module, generating attention maps that capture long-range dependencies between regions in the same window.         
        To enhance interpretability, we replaced the FCL with a convolutional classifier, referred to as the \emph{class evidence layer}. This layer leverages spatial information to produce class-wise evidence maps (Fig.\,\ref{fig1:architecture}), where each pixel reflects the local contribution of input regions to the final prediction. Following classification, the evidence maps are upsampled and overlaid on the input image for visualization purposes (Fig.\,\ref{fig1:architecture}d). 
        Furthermore, the inclusion of an explicit class evidence layer enables the application of an $\ell_1$ sparsity constraint on the class evidence maps $\mathbf{A}_c$, thereby enhancing interpretability \cite{donteu2023sparse, djoumessi2025soft}. This leads to the following loss function:  
        \begin{equation}
            \mathcal{L}(\mathbf{y},\mathbf{\hat{y}}) = \mbox{CE}(\mathbf{y}, \mathbf{\hat{y}}) + \lambda \sum_{i,j,c} |\mathbf{A}_c^{ij}| .
            \label{eq:loss}
        \end{equation}
        Here, CE denotes the cross-entropy loss, and $\mathbf{y}$ represents the reference class labels. The sparsity of the evidence maps is controlled by the hyperparameter $\lambda$. The entire model is trained end-to-end using gradient descent. 

\section{Results}
    \subsection{Datasets}
        We used two publicly available retinal fundus datasets, the Kaggle Diabetic Retinopathy (DR) \cite{kaggle_dr_detection} and the Age-Related Eye Disease Study (AREDS) \cite{study1999age}. 
        The Kaggle DR dataset had 45,923 images from 28,984 subjects after applying a custom quality filtering with class distributions: 73\% No DR, 15\% Mild, 8\% Moderate, 3\% Severe, and 1\% Proliferative DR. 
        The AREDS dataset contained 34,079 images from 4,757 participants. AMD severity was grouped into six categories \cite{al2017recent, study1999age}: 49\%, 19\%, 14\%, 3\%, 12\%, 1\% for early, moderate, advanced intermediate, early late, active neovascular, and end-stage AMD. 

        Images were resized to 512 × 512, normalized, and augmented with cropping, flipping, color jitter, and rotation. Datasets were split into 75\% training, 10\% validation, and 15\% test, keeping each participant’s records in the same split.
        We evaluated our model’s ability to localize DR-related lesions against ground-truth human annotations using the IDRiD dataset \cite{porwal2018indian}, which provides 81 fundus images with pixel-level labels for microaneurysms (MA), hemorrhages (HE), soft exudates (SE), and hard exudates (EX). This enabled the assessment of interpretability through localization performance.

        \begin{table}[b]
            \centering
            \caption{Classification performance on the test sets. Reported computational costs include: parameters (M), memory (MB), and average inference time (s). }
            \label{tab1}
            \begin{tabular}{l |c c c| c c | c c}
                \hline
                & \multicolumn{3}{c|}{Computational Cost} & \multicolumn{2}{c|}{AREDS AMD} & \multicolumn{2}{c}{Kaggle DR} \\
                 & \textbf{Par.} & \textbf{Mem.} & \textbf{Time} & \textbf{Acc.} & \textbf{$\kappa$} &  \textbf{Acc.} & \textbf{$\kappa$}  \\
                \hline
                ViT \cite{dosovitskiy2020image} & $86,094$ & $341$ & $09.5 \pm 0.1$ & $.76 \pm .03$ & $.90 \pm .02$ & $.81 \pm .02$ & $.71 \pm .04$ \\
                Swin \cite{liu2021swin} & $86,883$ & $358$ & $15.5 \pm 1.1$ & $.78 \pm 0.2$ & $\mathbf{.92 \pm .02}$ & $.85 \pm .02$ & $.81 \pm .03$ \\
                ResNet \cite{he2016deep} & $23,518$ & $101$ & $04.2 \pm 0.5$ & $.78 \pm .03$ & $.89 \pm .02$ & $.85 \pm .02$ & $.81 \pm .03$ \\
                BagNet \cite{donteu2023sparse} & $16,271$ & $193$ & $15.1 \pm 0.1$ & $.75 \pm .03$ & $.88 \pm .02$ & $.86 \pm .02$ & $.83 \pm.03$ \\
                \hline
                ResNet-FCL-SA & $69,732$ & $281$ & $06.2 \pm 0.2$ & $.78 \pm .03$ & $.90 \pm .02$ & $.86 \pm .02$ & $.82 \pm .03$ \\
                BagNet-FCL-SA & $62,501$ & $306$ & $27.3 \pm 0.2$ & $.77 \pm .03$ & $.89 \pm .02$ & $.85 \pm .02$ & $.83 \pm .03$ \\
                \hline
                ResNet-Conv-SA & $69,735$ & $285$ & $06.3 \pm 0.6$ & $.78 \pm .03$ & $.91 \pm .02$ & $.85 \pm .02$ & $.83 \pm .03$  \\
                BagNet-Conv-SA & $62,913$ & $310$ & $27.3 \pm 0.3$ & $.77 \pm.03$ & $.90 \pm .02$ & $\mathbf{.87 \pm .02}$ & $\mathbf{.84 \pm .02}$ \\
                \hline
                sResNet-Conv-SA & $69,735$ & $285$ & $06.3 \pm 0.6$ & $\mathbf{.79 \pm .02}$ & $.90 \pm .02$ & $.85 \pm .02$ & $.80 \pm .03$  \\
                sBagNet-Conv-SA & $62,913$ & $310$ & $ 27.3 \pm 0.3$ & $.77 \pm .03$ & $.91 \pm .02$ & $.85 \pm .02$ & $.81 \pm .03$ 
            \end{tabular}
            \label{tab:classification_results}
        \end{table} 

    \subsection{Self-explanaible hybrid models achieved SOTA performance} 
        We first evaluated our models on multiclass DR detection and AMD severity classification. Using the ResNet50 and BagNet-33 as backbone, our model incorporated dual-resolution convolutional self-attention (DR-Conv-SA) and a GDFN module \cite{zamir2022restormer}. We set the reduction factor to $r=2$, following \cite{ilyas2024hybrid}, and applied max pooling. The window size was tuned ($w=10$ for BagNet, $w=8$ for ResNet), along with the regularization coefficient $\lambda$ (Eq.~\ref{eq:loss}), to balance classification accuracy and evidence map sparsity. Classification metrics are reported with 95\% confidence interval (CI) lengths from bootstrapping, while inference time is reported with standard deviation (SD) over 1,000 runs on the same input.
        
        We compared our sparse models to the dense counterparts ($\lambda = 0$), a variant using linear self-attention (SA) with fully connected layer (FCL) classifier, and several other baselines: ResNet50, BagNet33, ViT32 (input size 384), and Swin Transformer (input size 384, patch size 4, window size 12). 
        All models were initialized with pre-trained weights from ImageNet and trained with the same setup: data augmentation, cross-entropy loss, cosine learning rate schedule, and SGD optimizer (learning rate $10^{-4}$, weight decay $5\times 10^{-4}$) 
        on an NVIDIA A40 GPU, using PyTorch, with model selection based on the best validation accuracy. 

        Our interpretable-by-design hybrid models achieved state-of-the-art performance on both tasks. The dense model with BagNet backbone yielded the best results for DR classification, while the model with the ResNet backbone achieved the highest Cohen’s kappa ($\kappa$) for AMD severity classification (Tab.\ref{tab:classification_results}). 
        Despite the sparsity penalty on the class activation map, the sparse models maintained competitive accuracy, with only a slight reduction in $\kappa$. Notably, for AMD detection, $\kappa$ exceeded accuracy, likely due to misclassifications occurring predominantly between adjacent severity levels (Fig.\,\ref{fig2:quantitative-eval}d). Overall, the computational cost varies depending on the backbone but remains lower than that of ViTs.        

    \subsection{Sparsity constraints enhance class evidence maps}
        \label{sec:qualitative_visualization}
        \begin{figure}[t]
            \centering
            \includegraphics[width=\textwidth]{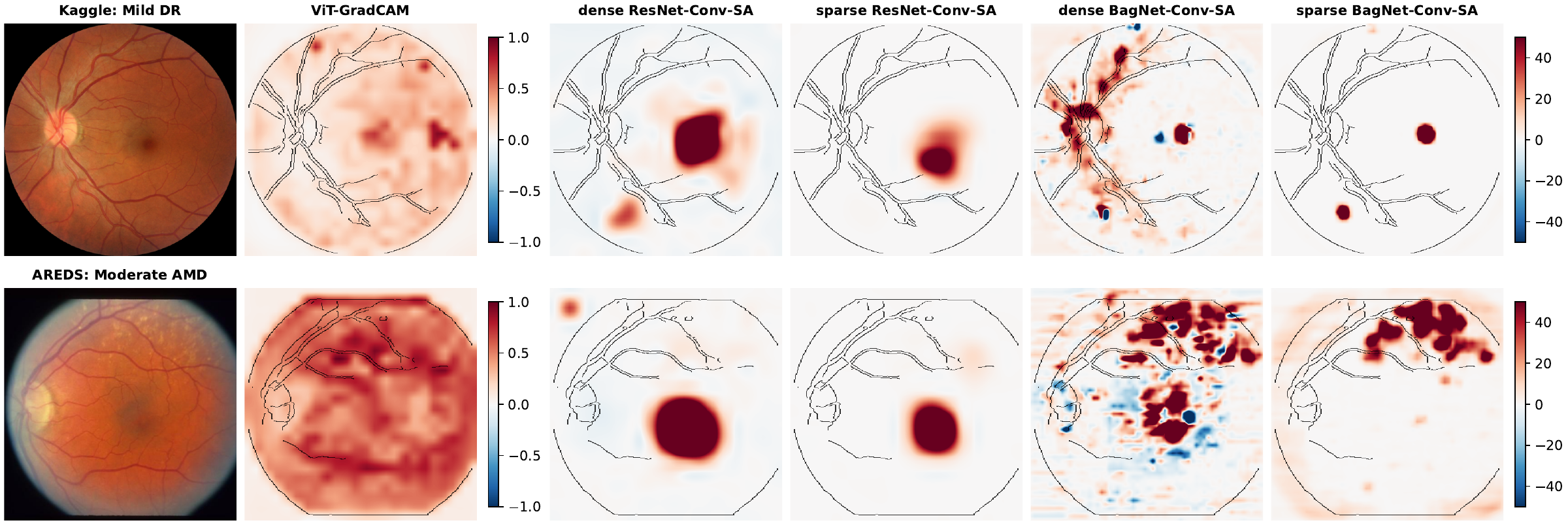}
            \caption{\textbf{Examples explanations}. From left to right, heatmaps for the correctly predicted class. The first row shows an example (grade 1) from the Kaggle dataset, while the second row shows an example (grade 2) from the AREDS dataset.}
            \label{fig2:visualization}
        \end{figure}
        
        We next compared evidence maps from our model to attribution maps generated with GradCAM \cite{selvaraju2017grad} on the ViT baseline. As these were multiclass tasks, we only showed class evidence maps from the correctly predicted class. Our class evidence maps, obtained from the convolutional layer before average pooling, clearly highlighted input features relevant to the predicted class (Fig.\,\ref{fig2:visualization}). 
        We noticed that GradCAM on ViT produced cluttered, hard-to-interpret heatmaps. In contrast, the hybrid ResNet-Transformer generated coarser heatmaps due to its large receptive field, while the hybrid BagNet-Transformer provided more localized explanations. The sparse models further refined this by producing sparser heatmaps, focusing decisions on smaller yet relevant retinal regions. For AMD severity classification, we observed that both the dense and sparse ResNet-Transformer models focus mainly on the macular region.

    \subsection{Evidence maps provide faithful and localized explanations}    
        \begin{figure}[t]
            \centering
            \includegraphics[width=\textwidth]{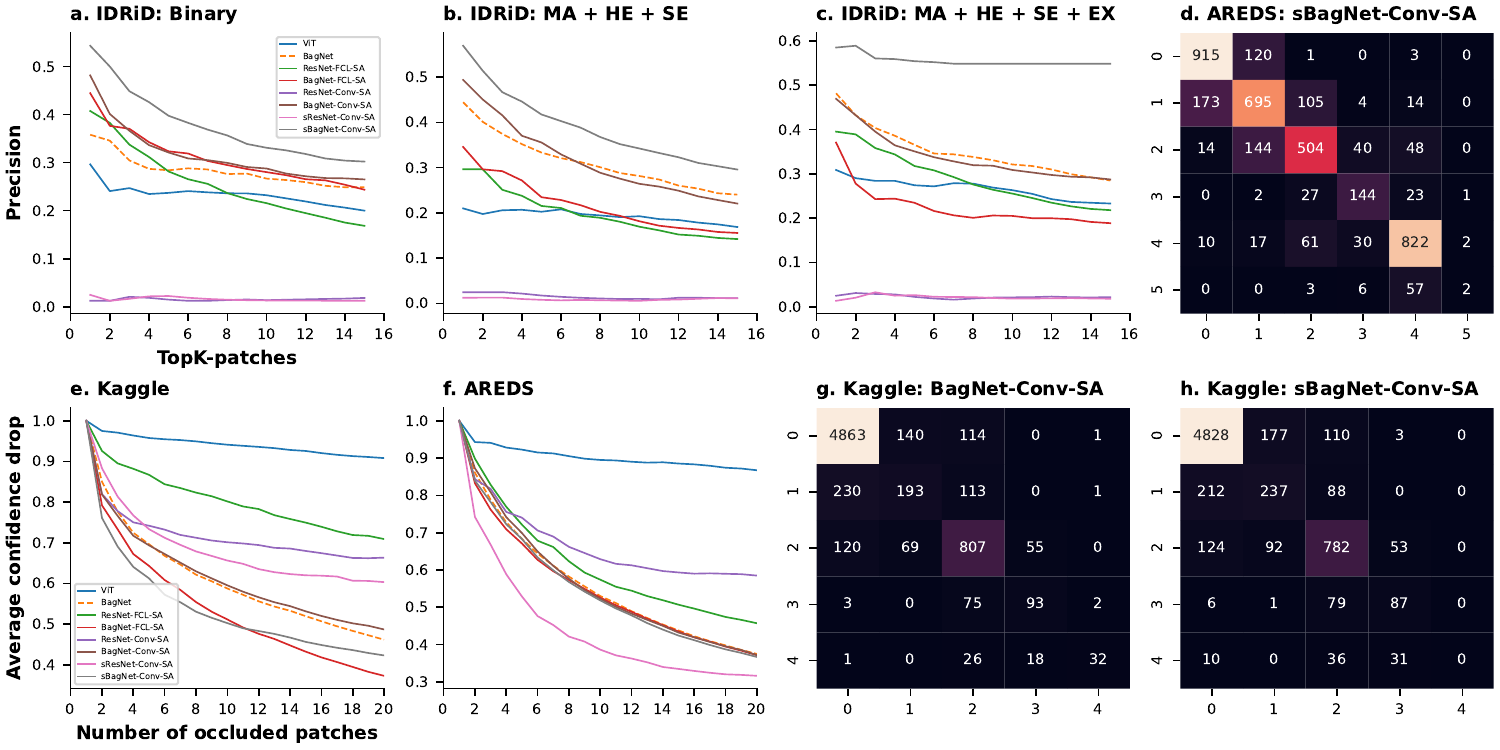}
            \caption{\textbf{Quantitative evaluation of heatmaps and confusion matrices.} (\textbf{a-c}) Precision evaluation on IDRiD dataset. (\textbf{e,f}) Sensitivity analysis of different heatmaps for DR detection and AMD severity classification. (\textbf{d,g,h}) Confusion matrices of different models for DR detection and AMD severity classification on the test sets.}
            \label{fig2:quantitative-eval}
        \end{figure}
        
        We quantitatively assessed the alignment of the explanations with clinical lesion-wise ground truth annotations by evaluating their precision in identifying DR lesions. 
        Following the International Clinical Diabetic Retinopathy Scale \cite{wilkinson2003proposed}, we evaluated three cases: (a) binary evaluation (Fig.\,\ref{fig2:quantitative-eval}a), averaging disease-class heatmaps and combining all lesion annotations; (b) severe DR  (Fig.\,\ref{fig2:quantitative-eval}b), where MAs, HEs, and SEs were combined, and the precision was computed from the severe grade heatmap (c) proliferative DR (Fig.\,\ref{fig2:quantitative-eval}c), where all lesions were combined and precision was evaluated from the heatmap from the proliferative grade heatmap. 
        Precision was measured as the proportion of positively activated regions containing lesions \cite{donteu2023sparse}, using $33 \times 33$ non-overlapping patches to match BagNet's receptive field. For ViT and hybrid FCL models, GradCAM-generated heatmaps were used. Patches were extracted from positively activated regions. 
        In all cases, the sparse BagNet-Transformer showed considerably higher precision than all other models and outperformed the base BagNet, suggesting that incorporating attention improved both classification and interpretability. The ResNet-Transformer with an explicit class-evidence layer performed worse, likely due to its larger receptive field producing coarser localizations (Fig.\,\ref{fig2:visualization}). 
        
        Subsequently, we additionally measured the faithfulness of the explanations by evaluating their ability to identify relevant regions for classification \cite{yeh2019fidelity}. Using correctly classified test images, we progressively removed top-ranked patches highlighted in the heatmap and measured the resulting drop in class confidence. 
        For DR detection, the sparse BagNet-Transformer performed best, while standard ViTs performed worst, followed by the ResNet-Transformer (Fig.\,\ref{fig2:quantitative-eval}e). 
        In contrast, for AMD severity classification, the hybrid sparse ResNet outperformed the sparse BagNet-Transformer (Fig.\,\ref{fig2:quantitative-eval}f), likely due to the larger lesion sizes in AMD, which favor CNNs with larger receptive fields. 
        Notably,  this trend was consistent with classification results, where the ResNet backbone also excelled.

    \subsection{Our model enhances interpretability for multi-class tasks}   
        \begin{figure}[t]
            \centering
            \includegraphics[width=\textwidth]{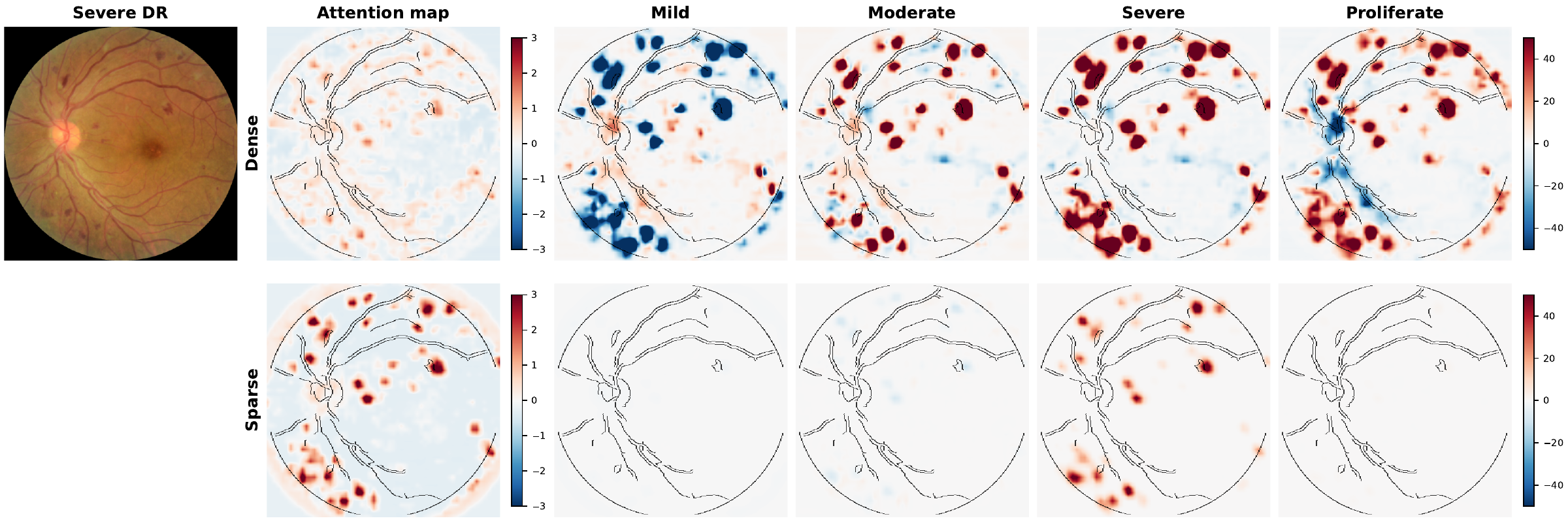}
            \caption{\textbf{Examples of multi-class explanations}. Class-specific heatmaps for a Severe DR example from the Kaggle dataset. The first row displays the attention map and corresponding heatmaps from the dense hybrid model with the BagNet backbone, while the second row shows the attention map and heatmaps from its sparse version.}
            \label{fig3:multiclass}
        \end{figure}
        
        Finally, we visualized class-specific explanations for the dense and sparse BagNet-Transformer. For DR prediction on the Kaggle dataset, both models correctly classified the example (Fig.\,\ref{fig3:multiclass}). 
        For our hybrid model, heatmaps and class probability distributions were generated in a single forward pass, with the sparse model producing more focused and localized explanations aligned with both the predicted class and clinical ground truth. In contrast, post-hoc explanations required multiple forward passes, increasing the overall inference cost. 
        In other classes, the sparse model showed almost no positive activations, unlike the dense model, which presented a mix of positive and negative evidence.
        Interestingly, we observed a strong correlation between attention maps and predicted evidence maps, particularly in the sparse model. This suggests that the model effectively captures long-range dependencies in an interpretable way.

\section{Discussion and Conclusion}
    We introduced the first inherently interpretable hybrid CNN-transformer architecture for medical image classification\footnote{Code at \url{https://github.com/kdjoumessi/Self-Explainable-CNN-Transformer}}, applied to DR detection and AMD severity classification from retinal fundus images. The approach is backbone-agnostic, allowing backbone selection to be guided by disease-specific prior.
    We evaluated the model with two CNN backbones: ResNet50, which captured global spatial relationships relevant to AMD, and BagNet, which aggregates small local features important for DR detection. 
    The latter is particularly noteworthy, as the SA mechanism helps overcome BagNet's limited receptive field. Conversely, ResNet’s larger receptive field is better suited for AMD, which involves larger lesions. In both cases, SA enhances the model's focus on the most relevant features.
    Our transformer module employs dual-resolution convolutional SA to capture both global and fine-grained features while preserving strong local inductive biases.
    Unlike standard models with FCL classifiers, our model includes an explicit class evidence layer that produces spatial class-evidence heatmaps, enabling direct interpretability without post-hoc methods. 

    Interestingly, the interpretability–accuracy trade-off was relatively small, challenging the myth of the accuracy-interpretability tradeoff in self-explainable models \cite{rudin2019stop}. All evaluated models achieved comparable performance, with high balanced accuracy and $\kappa$. Notably, the sparse BagNet-Transformer produced the most informative explanations for DR detection, while the sparse ResNet-Transformer yielded the best explanations for AMD severity classification. 
    
    Preliminary experiments showed that multi-head SA increased training time without improving classification, while multi-scale resolution had limited impacts and further increased both the training and inference time---particularly with the BagNet backbone (Tab.\,\ref{tab:classification_results}), due to its larger feature maps and the resulting higher SA computation cost. 
    Following \cite{donteu2023sparse}, we also observed that higher sparsity often led to missed detection of late-stage DR, likely due to their underrepresentation in the training set (Fig.\,\ref{fig3:multiclass}h). 
    However, our hybrid architecture mitigated this issue more effectively, demonstrating robustness in low-data settings. 
    Overall, our findings underscore hybrid CNN-Transformer models as a strong alternative to post-hoc ViT explanations, particularly for medical imaging.

    

    \begin{credits}
    \subsubsection{\ackname} This project was supported by the Hertie Foundation, the German Science Foundation (Excellence Cluster EXC 2064 ``Machine Learning—New Perspectives for Science'', project number 390727645). The authors thank the International Max Planck Research School for Intelligent Systems (IMPRS-IS) for supporting KD. PB is a member of the Else-Kröner-Kolleg ``ClinBrAIn''.

    \subsubsection{\discintname}
        The authors declare no competing interests.
    \end{credits}

    

\bibliographystyle{splncs04}
\bibliography{references}
\end{document}